\documentclass[conference]{IEEEtran}

\usepackage{amsfonts,amsmath,amsthm,amssymb}
\usepackage{enumerate}
\usepackage{graphics,graphicx,subfigure}

\usepackage{color}
\usepackage{todonotes}
\usepackage{cancel}
\usepackage{url}
\usepackage{hyperref}
\usepackage{makeidx}
\usepackage{showidx}
\usepackage{multicol}        
\usepackage{xspace}
\usepackage{stmaryrd}        
\usepackage{pifont}          
\usepackage{fancybox}        
\usepackage{bm}

\theoremstyle{plain}
 \theoremstyle{definition}
 \newtheorem{lem}{Lemma}
 \newtheorem{defn}[lem]{Definition}
 \newtheorem{thm}[lem]{Theorem}
 \newtheorem{prop}[lem]{Proposition}
 \newtheorem{cor}[lem]{Corollary}
 \newtheorem{notn}[lem]{Notations}
 \newtheorem{pb}[lem]{Problem}
 \newtheorem{form}[lem]{Formulae}
 
 \newtheorem*{rk}{Remark}
 \newtheorem*{com}{Comment}
 \newtheorem*{ex}{Example}
 \theoremstyle{remark}

 \newcommand{\blem}{\begin{lem}}
 \newcommand{\elem}{\end{lem}}
 \newcommand{\bdefn}{\begin{defn}}
 \newcommand{\edefn}{\end{defn}}
 \newcommand{\bthm}{\begin{thm} }
 \newcommand{\ethm}{\end{thm}}
 \newcommand{\bprop}{\begin{prop}}
 \newcommand{\eprop}{\end{prop}}
 \newcommand{\bcor}{\begin{cor}}
 \newcommand{\ecor}{\end{cor}}
 \newcommand{\bnotn}{\begin{notn}}
 \newcommand{\enotn}{\end{notn}}
 \newcommand{\bpb}{\begin{pb}}
 \newcommand{\epb}{\end{pb}}
 \newcommand{\bform}{\begin{form}}
 \newcommand{\eform}{\end{form}}
 \newcommand{\brk}{\begin{rk}}
 \newcommand{\erk}{\end{rk}}
 \newcommand{\bcom}{\begin{com}}
 \newcommand{\ecom}{\end{com}}
 \newcommand{\bex}{\begin{ex}}
 \newcommand{\eex}{\end{ex}}
 \newcommand{\bpf}{\begin{proof}}
 \newcommand{\epf}{\end{proof}}

\newcommand{\cE}{\mathcal{E}}

\newcommand{\cK}{\mathcal{K}}

\newcommand{\cS}{\mathcal{S}}

\newcommand{\bR}{\mathbb{R}}

\newcommand{\be}{\begin{equation}}
\newcommand{\ee}{\end{equation}}
\newcommand{\bal}{\begin{align}}
\newcommand{\eal}{\end{align}}
\newcommand{\ba}{\begin{align*}}
\newcommand{\ea}{\end{align*}}
\newcommand{\bmx}{\begin{matrix}}
\newcommand{\emx}{\end{matrix}}
\newcommand{\bbmx}{\begin{bmatrix}}
\newcommand{\ebmx}{\end{bmatrix}}
\newcommand{\bpmx}{\begin{pmatrix}}
\newcommand{\epmx}{\end{pmatrix}}
\newcommand{\bvmx}{\begin{vmatrix}}
\newcommand{\evmx}{\end{vmatrix}}

\newcommand{\wh}{\widehat}
\newcommand{\wt}{\widetilde}
\newcommand{\f}{\frac}
\newcommand{\df}{\dfrac}

\newcommand{\setm}{\setminus}

\newcommand{\argmin}{{\rm argmin}\,}

\newcommand{\minimize}[1]{\underset{#1}{\rm minimize}\,}

\newcommand{\la}{\lambda}
\newcommand{\La}{\Lambda}
\newcommand{\eps}{\varepsilon}
  
\usepackage{natbib}
\usepackage{censor}

\title{On the Optimal Recovery of Graph Signals}
\author{Simon Foucart, Chunyang Liao, and Nate Veldt --- Texas A\&M University}

\begin{document}

\maketitle

\begin{abstract}
Learning a smooth graph signal from partially observed data is a well-studied task in graph-based machine learning. 
We consider this task from the perspective of optimal recovery, a mathematical framework for learning a function from observational data 
that adopts a worst-case perspective tied to model assumptions on the function to be learned. 
Earlier work in the optimal recovery literature has shown that minimizing a regularized objective produces optimal solutions for a general class of problems, 
but did not fully identify the regularization parameter.
Our main contribution provides a way to compute
regularization parameters that are 
optimal or near-optimal (depending on the setting), 
specifically for graph signal processing problems.  
Our results offer a new interpretation for classical optimization techniques in graph-based learning
and also come with new insights for hyperparameter selection.
We illustrate the potential of our methods in numerical experiments on several semi-synthetic graph signal processing datasets.

\end{abstract}

\section{Introduction}

In graph signal processing, one starts with a dataset defined over an irregular graph domain 
and the goal is to recover a signal on vertices of the graph 
(e.g. discrete labels or regression values)
when given access to only one part of the signal~\citep{dong2020graph,ortega2018graph}. 
As a concrete example, the graph may encode US counties (nodes) and their physical adjacencies (edges) 
while the signals may represent voting patterns, birthrates, or any number of other attributes influenced by geographic region~\citep{jia2020residual}. 
As other examples,
in biology, the graph may represent a gene interaction network
while signals may indicate expression levels of individual genes~\citep{dong2020graph}; 
in neuroscience, brain activity signals coming from fMRI data may be analyzed over a graph representing physical connections or co-activations among regions of a brain~\citep{huang2018graph}, etc.	

The task of recovering a graph signal from partial information about it is also known as graph-based semi-supervised learning~\citep{zhou2003learning,belkin2004regularization,zhu2003semi}. 
This task has been studied in depth by researchers from many related academic communities including machine learning, statistics, and of course signal processing. 
In all of these settings, a common assumption is that signals vary smoothly over the graph's edge structure, 
meaning that adjacent nodes often share 
similar labels~\citep{zhou2003learning,zhu2003semi,belkin2004regularization,xu2010empirical,dong2020graph}. 
Many formal objective functions and theoretical results for graph signal processing and semi-supervised learning are justified by assuming that graph signals come from a certain well-behaved probability distributions~\citep{zhu2003semi,dong2019graph}. 
This often leads to objective functions that can be minimized using simple matrix-based methods~\citep{zhu2003semi,zhou2003learning,belkin2004regularization}.
However, the performance is affected by the choice of a regularization parameter in the objective function
and it is not always clear how to select such a parameter. 
In another direction, there has been a recent surge of interest in using graph-neural networks for learning over graphs.
This is often successful in practice but typically comes with no mathematical guarantees. 

In our work,
we address the graph signal processing task from a novel perspective---that of optimal recovery. 
This perspective does not rely on the assumption that ground truth signals are drawn from a well-behaved distribution. 
Instead, the goal is to find optimal solutions under worst-case assumptions about graph smoothness and labeling error. 
This approach comes with several benefits. 
Primarily, we present new theoretical results on finding best solutions under the optimal recovery framework 
(locally and globally, see later sections for technical details). 
Along the way, we highlight the connections between the optimization problems stemming from this framework and the classical techniques encountered in graph signal processing. 
One significant contribution is to provide rigorous theoretical guarantees for selecting the regularization parameter in the objective function being minimized. 
Setting this parameter is not entirely free of challenges, 
as it actually depends on the parameters characterizing graph smoothness and labeling error. 
Nevertheless, our results offer fresh intuition 
on how to reasonably choose the regularization parameters intrinsic to objective functions common in graph signal processing. 
Finally, we provide a proof-of-concept implementation of our approach and illustrate its performance in several empirical graph signal processing experiments.

\section{The Perspective from Optimal Recovery}

Let $G=(V,E)$ be an undirected graph with $N=|V|$ vertices identified with $1,2,\ldots,N$.
A signal $f$ defined on $V$ is thus identified with a vector $f \in \bR^N$.
The previously-mentioned common assumption that  $f$ varies smoothly over the graph's edge structure qualitatively translates into the fact that the values $f_i$ and $f_j$ do not differ much if the vertices $i$ and $j$ are strongly connected.
Quantitatively,
putting a weight $w_{i,j} \ge 0$ on the edge connecting $i$ and $j$, thus defining a (weighted, symmetric) adjacency matrix $W \in  \bR^{N \times N}$, the assumption takes the form
$$
\f{1}{2} \sum_{i,j=1}^N w_{i,j}(f_i - f_j)^2 \le \eps^2
$$
for a small $\eps > 0$ standing for a graph smoothness parameter.
Introducing the graph Laplacian $L = D-W \in \bR^{N \times N}$,
where $D$ is the diagonal matrix with entries $D_{i,i} = \sum_{j =1}^N W_{i,j}$,
the assumption succinctly reads
$$
\langle Lf, f \rangle  \le \eps^2,
\qquad \mbox{or} \quad
\| L^{1/2} f \|_2 \le \eps.
$$
We recall that the square-root $L^{1/2}$ of $L$ is well-defined because the graph Laplacian $L$ is positive semidefinite.
Note that it is not positive definite,
since $0$ is always an eigenvalue of $L$.
In fact, its multiplicity equals the number of connected components $C$ of $G$,
with orthogonal eigenvectors provided by the indicator vectors ${\bm 1}_C \in \{0,1\}^N$ of $C$.
Throughout, we shall assume that the graph $G$ is known,
and hence that $L$ is available to the user.

As for the unknown $f$,
it is partially observed---or labeled\footnote{Labels are real numbers here, not elements of a binary set such as $\{0,1\}$.}.
In other words,
there is a subset $V_\ell$, with size $|V_\ell | = n_\ell$, of vertices for which the $f_i$, $i \in V_\ell$,
are known.
In reality, they are known up to additive errors,
so that the user has access to
$$
    y_i = f_i + e_i,
    \qquad i \in V_\ell.
$$
To abbreviate, we write $y = \La f + e \in \bR^{n_\ell}$,
where the linear map $\La: \bR^N \to \bR^{n_\ell}$ satisfies $\La \La^* = I_{n_\ell}$ here
(since, up to a proper ordering of the vertices, (the matrix of)  $\La$ takes the form $\begin{bmatrix} \; I_{n_\ell} \; | \; 0 \; \end{bmatrix}$).
We shall assume that an $\ell_2$-bound on the error vector $e$ is available, 
namely 
$$
\|e\|_2 \le \eta
$$
for a small $\eta > 0$ standing for a labeling error parameter.

Our objective is now to estimate the graph signal $f$ on the set of unlabeled vertices, 
i.e., on $V_u = V \setm V_\ell$,
which has size $|V_u| = n_u = N - n_\ell$.
In the framework of optimal recovery, 
we aim at doing so in a worst-case optimal way 
given the graph smoothness and labeling error assumptions,
expressed as $f \in \cK$ and $e \in \cE$, where
the model set $\cK$ and the uncertainty set $\cE$ are given as
\begin{align}
\label{Model}
    \cK & = \{ f \in \bR^N: \|L^{1/2} f \|_2 \le \eps \},\\
\label{Uncertainty}    
    \cE & = \{ e \in \bR^{n_\ell}: \|e\|_2 \le \eta \}.
\end{align}

A scheme to estimate $f_{|V_{u}} \in \bR^{n_u}$ from $y \in \bR^{n_\ell}$ is nothing but a map $\Delta: \bR^{n_\ell} \to \bR^{n_u}$,
which we call a recovery map.
We are interested in those recovery maps that are optimal 
\begin{itemize}
\item {\em in the global setting,} 
i.e., 
$$
\sup\{ \|f_{|V_u} - \Delta(\La f  + e) \|_2: f \in \cK, e \in \cE \}
$$
is as small as possible;
\item {\em in the local setting},
i.e.,
at any given $y \in \bR^{n_\ell}$,
$$
\sup\{\|f_{|V_u} - z \|_2: f \in \cK, e \in \cE, \La f + e = y\}
$$
evaluated at $z = \Delta(y)$
is as small as possible.
Such a $\Delta(y) \in \bR^{n_u}$ is called a Chebyshev center for the set $\cS = \{ f_{|V_u}: f \in \cK, e \in \cE, \La f + e = y \}$, 
as it is easily seen to be a center of a minimal-radius ball containing~$\cS$.
\end{itemize}
If we believe that the observed labels need to be adjusted, too,
instead of estimating $f_{|V_u}$,
we may want to estimate $f$ in full.
We may also want to estimate the average of $f$
or its value at a particular vertex $i_0 \in V$.
To deal with these situations all at once,
we introduce a quantity of interest $Q: \bR^N \to \bR^n$,
which in the examples above is given by, respectively,
$$
Q(f) = f_{|V_u},
\; \;
Q(f) = f,
\; \; 
Q(f) = \f{1}{N} \sum_{i \in V} f_i,
\; \;
Q(f) = f_{i_0}.
$$
In this generality,
the global and local worst-case errors for the estimation of $Q$
are defined,
for $\Delta: \bR^{n_\ell} \to \bR^n$
and for $y \in \bR^{n_\ell}$, $z \in \bR^n$,
by
\begin{align*}
{\rm gwce}_Q(\Delta) & = \sup_{f \in \cK, e \in \cE} \{ \|Q(f) - \Delta(\La f  + e) \|_2 \},\\
    {\rm lwce}_Q(y,z) & =  
    \sup_{f \in \cK, e \in \cE} \{\|Q(f) - z \|_2:  \La f + e = y\}.
\end{align*}
We call a recovery map $\Delta: \bR^{n_\ell} \to \bR^n$
globally optimal if it minimizes ${\rm gwce}_Q(\Delta)$
and locally optimal if $\Delta(y)$ minimizes ${\rm lwce}_Q(y,z)$ for any given $y \in \bR^{n_\ell}$.
Of course,  locally optimal recovery maps are automatically globally optimal,  but they are typically harder to produce 
(as the current work will also illustrate). 
We may therefore relax the aspiration of genuine optimality to one of near optimality by merely requiring that ${\rm lwce}_Q(y,\Delta(y)) \le C \, \inf\{ {\rm lwce}_Q(y,z), z \in \bR^n \}$ for some absolute constant $C>1$.

\section{Selection of the Regularization Parameter}
We now show that (near-)optimal recovery maps can be obtained through Tikhonov-style regularization 
and we uncover a principled way to choose the regularization parameter based on the graph smoothness and labeling error parameters.
We start with some preparatory information about regularization 
before presenting our genuine optimality result in the global setting and our near optimality result in the local setting.

\subsection{Rundown on regularization}

When searching for the signal $f \in \bR^N$ that produced the observation vector $y \in \bR^{n_\ell}$,
it is natural to try and make the data-fidelity term $\|\La f - y\|_2^2$ small.
Furthermore,
to enforce the graph smoothness condition that $\| L^{1/2} f\|_2^2$ is small,
one can incorporate this condition as a constraint in a miminization problem or add the regularization term $\gamma \|L^{1/2} f\|_2^2$ to the objective function,
as done e.g. in \citep{belkin2004regularization}.
Instead of parametrizing by $\gamma > 0$,
it will be more convenient for our purpose to parametrize by some $\tau \in (0,1)$,
thus leading to the regularization map $\Delta_\tau: \bR^{n_\ell} \to \bR^N$ given by
$$
\Delta_\tau : y \mapsto
\underset{f \in \bR^N}{\argmin} \; (1-\tau) \|L^{1/2} f\|_2^2 + \tau \|\La f - y \|_2^2 . 
$$
This map is well defined under the assumption that at least one vertex is observed in each connected component of the graph,
which translates into $\ker(L) \cap \ker(\La) = \{0\}$
or equivalently into the invertibility of $(1-\tau) L + \tau \La^* \La$  (see \S 1 in the supplementary material).
Indeed, as the minimizers $f_\tau$ of the above objective function are characterized by the normal equation $(1-\tau) L f_\tau + \tau \La^* (\La f_\tau - y) =0$,
this invertibity shows that $f_\tau$ is unique and is equal 
to 
\be
\label{ExprReg}
\Delta_\tau(y) = \big( (1-\tau) L + \tau \La^* \La \big)^{-1} (\tau \La^* y).
\ee
This expression reveals in particular that $\Delta_\tau$ is a linear map.

The extreme case $\tau \to 0$ is interpreted as the minimizer of $\|\La f -y\|_2^2$ subject to $L^{1/2} f =0$,
which is not very interesting, see \S 2 for explanation.
The extreme case $\tau \to 1$ is interpreted as the minimizer of $\|L^{1/2}f\|_2^2$ subject to $\La f = y$,
which appears commonly in graph signal processing under the names of harmonic method \citep{zhu2003semi} or  interpolatory method \citep{belkin2004regularization}.

\subsection{Genuine optimality in the global setting}

It was recognized already in 
\citep{micchelli1979inaccurate,micchelli1993optimal} that the regularization maps $\Delta_\tau$ produce a globally optimal recovery map for {\em some} parameter $\tau \in (0,1)$,
but the choice of this parameter was not made explicit.
Theorem~\ref{ThmGlobal} below shows that this parameter can be obtained by solving a semidefinite program.
Such a result was established in \citep{foucart2022optimal} in a slightly more restrictive setting,
namely the place of $L^{1/2}$ was taken by an orthogonal projector and only full recovery (i.e., $Q=I_N$) was considered. 

\bthm
\label{ThmGlobal}
Given a linear quantity of interest $Q: \bR^N \to \bR^n$, 
let $\cK$  and $\cE$ be the model and uncertainty sets from \eqref{Model}-\eqref{Uncertainty}.
Defining $\tau_\flat = d_\flat/(c_\flat + d_\flat)$
where $c_\flat,d_\flat \ge 0$ are solutions to
\be
\label{SDPGlobal}
\minimize{c,d \ge 0} \; c \eps^2 + d \eta^2 
\quad \mbox{s.to } \; c L + d \La^* \La \succeq Q^* Q,
\ee
the linear map $Q \circ \Delta_{\tau_\flat} : \bR^{n_\ell} \to \bR^n$
is a globally optimal recovery map relative to $\cK$ and $\cE$,
meaning that
$$
{\rm gwce}_Q(Q \circ \Delta_{\tau_\flat}) = \inf_{\Delta: \bR^{n^\ell} \to \bR^{n}} {\rm gwce}_Q(\Delta).
$$
\ethm

The justification of this theorem relies on the two lemmas below,
whose proofs appear in \S 3 and \S 4 of the supplementary material.
Lemma \ref{LemLBGlobal} 
relies on a version of the S-procedure due to \cite{polyak1998convexity}
and follows \citep{foucart2022optimal} closely,
but the argument for Lemma \ref{LB+UBGlobal} follows a different route to deal with a quantity of interest $Q \not= I_N$.

\blem
\label{LemLBGlobal}
For an arbitrary recovery map $\Delta: \bR^{n_\ell} \to \bR^n$,
the squared global worst-case error is lower-bounded as
\begin{align*}
{\rm gwce}_Q(\Delta)^2
& \ge \sup_{h \in \bR^N} \{ \|Q(h)\|_2^2:  \| L^{1/2} h \|_2^2 \le \eps^2,  \|\La h \|_2^2 \le \eta^2 \}\\
& = \inf_{c,d \ge 0} \{ c \eps^2 + d \eta^2: c L + d \La^* \La \succeq Q^* Q \}. 
\end{align*}
\elem

\blem
\label{LB+UBGlobal}
If $c,d \ge 0$ satisfy $c L + d \La^* \La \succeq Q^* Q$, then, setting $\tau = d/(c+d)$, 
one has, 
for all $f \in \bR^N$ and all $e \in \bR^{n_\ell}$,
$$
\|Q(I-\Delta_\tau \La)f - Q  \Delta_\tau e \|_2^2 \le c \|L^{1/2} f\|_2^2 + d \|e\|_2^2.
$$
\elem

\bpf[Proof of Theorem \ref{ThmGlobal}]
Let $c_\flat,d_\flat \ge 0$ be minimizers of \eqref{SDPGlobal}.
On~the one hand,
according to Lemma \ref{LemLBGlobal},
the squared global worst-case error of any recovery map $\Delta: \bR^{n_\ell} \to \bR^n$ satisfies
$$
{\rm gwce}_Q(\Delta)^2 \ge c_\flat \eps^2 + d_\flat \eta^2.
$$
On the other hand, 
the linearity of $Q \circ \Delta_{\tau_\flat}$, $\tau_\flat := d_\flat/(c_\flat + d_\flat)$,
implies that its squared global worst-case error  becomes
$$
{\rm gwce}_Q(Q \circ \Delta_{\tau_\flat})^2 
= \sup_{f \in \cK, e \in \cE}
\|Q(I-\Delta_\tau \La)f - Q  \Delta_\tau e \|_2^2,
$$
which, according to Lemma \ref{LB+UBGlobal}, does not exceed
$$ 
\sup_{f \in \cK, e \in \cE}
c \|L^{1/2} f\|_2^2 + d \|e\|_2^2
 = c_\flat \eps^2 + d_\flat \eta^2.
$$
All in all, we have shown that ${\rm gwce}_Q(Q \circ \Delta_{\tau_\flat}) \le {\rm gwce}_Q(\Delta)$
for any map $\Delta: \bR^{n_\ell} \to \bR^n$,
which is the desired result.
\epf

\brk
To achieve genuine optimality,
exact knowledge of the parameters $\eps$ and $\eta$ is needed,
but near optimality is achievable when these are overestimated by some $\bar{\eps}$ and $\bar{\eta}$ satisfying
$\bar{\eps} \le C \eps$ and $\bar{\eta} \le C \eta$, see \S 5. 
\erk

\brk
The semidefinite program \eqref{SDPGlobal}, featuring an $N \times N$ matrix,
does not run when $N$ is in the thousands.
Nonetheless, it is expected that the computational burden could be alleviated if $Q$ maps into a low-dimensional space~$\bR^n$.
This is certainly the case in the extreme case $n=1$,
see \S 6.
\erk

\subsection{Near optimality in the local setting}

In contrast to the global setting,
we are unaware of a general result stating that 
the regularization maps $\Delta_\tau$ produce a locally optimal recovery map for {\em some} parameter $\tau \in (0,1)$.
For full recovery ($Q=I_N$), such a statement is true in at least two situations, though.
The first situation 
requires $\La \La^* = I_{n_\ell}$
(which is the case here)
and an orthogonal projector $P$ in place of $L^{1/2}$
(up to normalization, $L^{1/2}$ happens to be an orthogonal projector if $G$ is an unweighted graph made of disconnected complete subgraphs of identical sizes, see \S 7):
it was established in \citep{foucart2022optimal} that $\Delta_{\tau_\sharp}$ is a locally optimal recovery map
when $\tau_\sharp$ is the unique parameter $\tau$ between $1/2$ and $\eps/(\eps+\eta)$ satisfying the eigenvalue equation
{\small
\begin{multline*}
\la_{\min} ((1-\tau)P + \tau \La^* \La)\\
= \f{(1-\tau)^2 \eps^2 - \tau^2 \eta^2}{(1-\tau) \eps^2 - \tau \eta^2 + (1-\tau) \tau (1-2 \tau)  \delta^2},
\end{multline*}
}where  $\delta \hspace{-1mm}=\hspace{-1mm} \min\{ \|P f\|_2 \hspace{-1mm} : \hspace{-1mm} \La f \hspace{-1mm}=\hspace{-1mm} y \} \hspace{-1mm}=\hspace{-1mm} \min \{ \|\La f - y\|_2 \hspace{-1mm} : \hspace{-1mm} Pf \hspace{-1mm} = \hspace{-1mm}0 \}$.
The~second situation requires working in the complex setting:
it was established in \citep{beck2007regularization} that 
$\Delta_{\tau_\flat}$ is a locally optimal recovery map
when $\tau_\flat = d_\flat/(c_\flat + d_\flat)$,
with $c_\flat,d_\flat$ solving the semidefinite program
\begin{align*}
\minimize{\substack{c,d, t \ge 0}} \; c \eps^2  + d (\eta^2  - & \|y\|_2^2) + t 
\quad \mbox{s.to }  c L + d \La^* \La \succeq I_N\\
& \mbox{and } 
\bbmx
c L + d \La^* \La & \vline & d \La^* y\\
\hline
d y^* \La & \vline & t
\ebmx \succeq 0.
\end{align*}
In the real setting, although the value of ${\rm lwce}_{I_N}(y,\Delta_{\tau_\flat}(y))$ is only guaranteed to provide an upper bound for the minimal local worst-case error,
it is not unlikely that $\Delta_{\tau_\flat}$ is genuinely a locally optimal recovery map---this is the case in the first situation (result not published yet). 

Regardless of the above considerations, 
relaxing genuine optimality to near optimality,
it can always be guaranteed that some regularization map $\Delta_\tau$ produces a locally near optimal recovery map for a parameter $\tau \in (0,1)$ that can be computed.
This is the gist of the following result,
see \S 8 for a proof.

\bthm
\label{ThmLocal}
Given a linear quantity of interest $Q: \bR^N \to \bR^n$, 
let $\cK$  and $\cE$ be the model and uncertainty sets from \eqref{Model}-\eqref{Uncertainty}.
For $y \in \bR^{n_\ell}$,
let $\wh{f} \in \bR^N$ be the solution to\vspace{-1mm}
$$
\underset{f \in \bR^N}{\rm minimize \;} \max\Big\{  
\|L^{1/2} f\|_2^2, \f{\eps^2}{\eta^2} \|\La f - y\|_2^2\Big\}.\vspace{-1mm}
$$
Then $\wh{f}$ agrees with $\Delta_{\tau_\natural}(y)$,
where $\tau_\natural$ is the unique parameter $\tau \in (0,1)$ satisfying\vspace{-1mm}
$$
\|L^{1/2} \Delta_\tau(y)\|_2 = \f{\eps}{\eta} \|\La  \Delta_\tau(y) - y \|_2.\vspace{-1mm}
$$
Moreover, one has
${\rm lwce}_Q(y,Q(\wh{f}))
\le 2 \inf_{z \in \bR^n} {\rm lwce}_Q(y,z)$.
\ethm

\section{Numerical Validation}

In this section, we illustrate the performance of optimal recovery methods on several semi-synthetic regression datasets and verify the near optimality of a regularization parameter in the local setting. We implement our algorithms in {\sc matlab} and use {\sf CVX} \citep{CVX} for solving semidefinite programs.  
All numerical experiments are available on the GitHub repository
\url{https://github.com/liaochunyang/ORofGraphSignals}.

Let us recall that all our optimal recovery maps correspond to solving a regularized optimization program with a specific choice of hyperparameter. 
This type of regularized objective is already standard in graph-based machine learning~\citep{belkin2004regularization},
but it is often unclear how this parameter should be chosen. 
The primary goal of our numerical experiments is to illustrate how our techniques for hyperparameter selection work in a controlled environment where we have access to a ground truth graph signal $f$ and we can estimate the true smoothness and error parameters $\epsilon$ and $\eta$. 
In practical settings, 
we do not actually have access to $f$, 
nor can we estimate $\eps$ and $\eta$ exactly, 
but we shall confirm that near optimality is achievable under mild overestimations of $\epsilon$ and $\eta$ (see \S 5). 

We consider several real-world graphs and,
using a standard approach~\citep{dong2019graph} (see \S 9 for more details),
we generate synthetic signals $f$ 
whose values at the nodes are normalizing to be between 0 and 1 for simplicity. 
In our first experiment,
we show how the prediction error changes as the number of labeled vertices grows. 
We begin by running all methods for $n_\ell = 5$
and we keep adding $5$ new labeled nodes at a time---by the end, we have run experiments for $n_\ell$ ranging from $5$ to $N/2$ in increments of $5$. 
For each choice of $n_\ell$, 
our goal is to recover labels at unlabeled nodes, i.e., $Q(f)=f|_{V_u}$,
where $f$ is the vector of true node labels.
The prediction error for any estimator $\hat{f}$ of $f|_{V_u}$ is defined as $\|f|_{V_u}-\hat{f}\|_2$. 
For the smoothness parameter,
we set $\epsilon = 2 \|L^{1/2} f\|_2^2$.
Next, we introduce noise artificially by generating a uniform random vector $e$ and subtract the mean before scaling so that $\|e\|_2 \leq \eta$, where $\eta$ is chosen as $\eta=2$ (see \S 9 for details on other types of noise). 
We then create the corrupted labels $y = f_{|V_{\ell}} + e$. 
In the supplementary material, we also consider a mild overestimation on $\eta$ by setting $\bar{\eta}=2\eta$.

\begin{figure}[!htbp]
\centering     
\includegraphics[width=60mm]{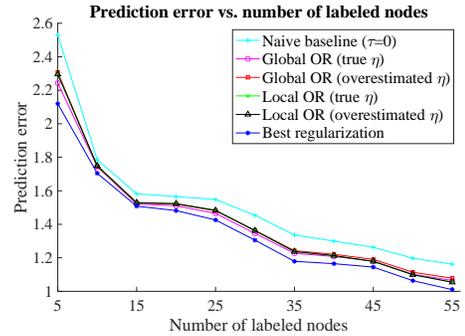}
\caption{Prediction error vs. number of labeled nodes on graph Adjnoun with added uniform noise.}
\label{prediction}
\end{figure}

In Figure \ref{prediction}, we display the prediction errors produced by locally/globally optimal recovery maps for the graph Adjnoun (see \S 9 for results on other graphs). 
The constructions of the globally optimal recovery map and the locally near optimal recovery map were presented in Theorem~\ref{ThmGlobal} and Theorem~\ref{ThmLocal}, respectively. 
We also use a grid-search approach to find the smallest prediction error over all regularization parameters in order to display a curve of the lowest possible prediction error. 
This is neither computationally efficient nor realistic, 
as it assumes that we can always check the prediction error for any estimator $\hat{f}$, 
but it provides a bound on the best case scenario for solving the regularized objective. 
Comparing the magenta and the red lines and the black and the green lines, 
we observe that an overestimation of $\eta$ does not lead to large differences in prediction error for both local and global optimal recovery maps, 
which suggests that one can safely use a mild overestimation of $\eta$ when we cannot access the true $\eta$. 
We also notice that the prediction errors produced by locally/globally optimal recovery maps are close to the prediction error produced by the best Tikhonov regularization method (blue line). 

\begin{figure}[!htbp]
\centering     
\includegraphics[width=60mm]{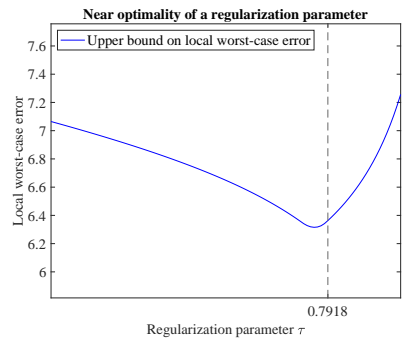}
\caption{Local worst-case error vs. regularization parameter.}
\label{local}
\end{figure}

In the second experiment (see Figure \ref{local}),
we confirm the near optimality of the locally optimal recovery map described in Theorem \ref{ThmLocal}. 
The setup of this experiment is similar to the first experiment. 
The parameter $\tau_\sharp$---the unique $\tau\in(0,1)$ such that $\|L^{1/2} \Delta_\tau(y)\|_2 = (\eps/\eta) \|\La  \Delta_\tau(y) - y \|_2$--- is found by the bisection method and is displayed in Figure \ref{local}
by the dashed vertical line.
The blue curve represents an upper bound for the local worst-case error ${\rm lwce}_Q(y,Q\circ\Delta_\tau(y))$
as a function of the regularization parameter.
For each $\tau$ in a grid of $[0,1]$,
this upper bound was computed by solving a semidefinite relaxation for the local worst-case error,
see \S 10 for details.
Figure \ref{local} not only supports the local near optimality of the recovery map $Q\circ\Delta_{\tau_\sharp}$,
but it also hints that $\tau_\sharp$ is not far away the best regularization parameter. 

\section*{Acknowledgment}

S. F. is partially supported by grants from the NSF (DMS-2053172) and from the ONR (N00014-20-1-2787).

\newpage

\bibliography{refs}

\newpage



\section*{Supplementary Material}

In this section, 
we supply the theoretical justifications that are missing from the main text.

\vskip 5mm

\paragraph*{{\bf \S 1.} The regularization map is well defined for $\tau \in (0,1)$}

The regularization problem can be written as
$$
\underset{f \in \bR^N}{\rm minimize \;} \|A f - b\|_2^2,
\quad 
A := \bbmx \sqrt{1-\tau} L^{1/2} \\ \hline \sqrt{\tau} \La \ebmx,
\; \; b := \bbmx 0 \\ \hline \sqrt{\tau} y \ebmx.
$$
Its solutions $\bar{f}$ are characterized by the normal equation $A^* A \bar{f} = A^* b$,
i.e., by $\big( (1-\tau) L + \tau \La^* \La \big) \bar{f} = \tau \La^* y$.
Note that we always make the assumption $\ker(L) \cap \ker(\La) = \{ 0 \}$,
otherwise,
fixing $f_0 \in \cK$ and $e_0 \in \cE$ with $\La f_0 + e_0 = y$,
the existence of $h \in \bR^N \setm \{ 0 \}$ such that $Lh = 0$  and $\La h = 0$
implies that $f_t := f_0 + t h \in \cK$, $e_t := e_0 \in \cE$, and $\La f_t + e_t = y$ for all $t \in \bR$,
yielding an infinite local (in turn global) worst-case error for the recovery of $Q = I_N$.
This assumption ensures that $(1-\tau) L + \tau \La^* \La $
is positive definite---hence invertible---for any $\tau \in (0,1)$,
since 
$$
\big\langle \big( (1-\tau) L + \tau \La^* \La \big) h,h \big\rangle = (1-\tau) \|L^{1/2}h\|_2^2 + \tau \|\La h \|_2^2 \ge 0  
$$
for all $h \in \bR^N$, with equality possible when and only when $h \in \ker(L) \cap \ker(\La) = \{ 0 \}$.
This shows that $\bar{f}$ is unique and given by 
$\bar{f} = \big( (1-\tau) L + \tau \La^* \La \big)^{-1} (\tau \La^* y)$.
Finally, if the graph $G$ is made of $K$ connected components $C_1,\ldots, C_K$, we observe that
\begin{align*}
\ker&(L)  \cap \ker(\La) 
 = \left\{ h = \sum_{k=1}^K a_k {\bm 1}_{C_k}, \, a \in \bR^K, \, h_{|V_\ell} = 0 \right\} \\
& = \left\{\sum_{k=1}^K a_k {\bm 1}_{C_k}, \, a \in \bR^K, \, a_k = 0 \mbox{ when } C_k \cap V_\ell \not= \emptyset \right\},
\end{align*}
 so that $\ker(L)  \cap \ker(\La)  = \{ 0\}$ if and only if $C_k \cap V_\ell \not= \emptyset$ for all $k = 1,\ldots,K$,
 which means that at least one vertex is observed in each connected component.

\vskip 5mm

\paragraph*{{\bf \S 2.} The limiting case $\tau \to 0$}

Writing $f_\tau = \Delta_\tau(y)$,
if we divide the objective function that $f_\tau$ minimizes by $\tau > 0$,
we see that
$$
f_\tau = \underset{f \in \bR^N}{\argmin \; }
\f{1-\tau}{\tau} \|L^{1/2} f\|_2^2 + \|\La f_\tau -y \|_2^2.
$$
Intuitively, the limit $f_0$ of $f_\tau$ as $\tau \to 0$ should satisfy $L^{1/2} f_0 = 0$,
otherwise $\|L^{1/2} f_\tau\|_2^2 \ge \kappa$ for some $\kappa > 0$ when $\tau$ is sufficiently small,
and then $\big( (1-\tau)/\tau \big) \|L^{1/2} f_\tau\|_2^2$ blows up as $\tau \to 0$,
preventing $f_\tau$ to be a minimizer of the divided objective function.
It suggests---and this can be made precise---that
$$
f_0 = \underset{f \in \bR^N}{\argmin \; }
\|\La f -y \|_2^2
\quad \mbox{s.to } \; L^{1/2}f=0.
$$
The constraint $L^{1/2}f=0$ is equivalent to $f$ taking the form $f= \sum_{k=1}^K a_k {\bm 1}_{C_k}$ for some $a \in \bR^K$,
where $C_1,\ldots,C_K$ denote the connected components of the graph $G$.
Under this constraint,
we then have
$$
\La f - y = f_{|V_\ell} - y = \sum_{k=1}^K \big( a_k {\bm 1}_{C_k \cap V_\ell} - y_{C_k} \big), 
$$
and hence, since the summands have disjoint supports,
\begin{align*}
\|\La  f - & y\|_2^2  
 = \sum_{k=1}^K \| a_k {\bm 1}_{C_k \cap V_\ell} - y_{C_k}  \|_2^2\\
& = \sum_{k=1}^K \big( a_k^2  |C_k \cap V_\ell| - 2 a_k \langle {\bm 1}_{C_k \cap V_\ell}, y_{C_k} \rangle + \|y_{C_k}\|_2^2 \big).
\end{align*}
This quantity is easily seen to attain its minimal value when $a_k = \langle {\bm 1}_{C_k \cap V_\ell}, y_{C_k} \rangle / |C_k \cap V_\ell|$ for each $k=1,\ldots,K$.
All in all, this signifies that the component of $f_0$ on each $C_k$ is equal to the average of $y_{C_k}$---as announced, the case $\tau \to 0$ is not very interesting!

\vskip 5mm

\paragraph*{{\bf \S 3.} Proof of Lemma \ref{LemLBGlobal}}
To prove the inequality,
consider $h\in\bR^N$ with  $\|L^{1/2 }h \|_2^2 \le \eps^2$ and $\| \La h\|_2^2 \le \eta$.
Then define $f_\pm = \pm h \in \cK$  and $e_\pm = \mp \La  h \in \cE$, so that 
\begin{align*}
{\rm gwce}_Q(\Delta) & \geq \max_\pm \| Q(f_\pm) - \Delta( \La f_\pm + e_\pm )\|_2 \\
& = \max_\pm \| Q(\pm h) - \Delta(0) \|_2  \\
& \geq \frac{1}{2} \|Q(h) - \Delta(0) \|_2 + \frac{1}{2} \|Q(-h) - \Delta(0) \|_2 \\
& \geq \frac{1}{2} \| (Q(h)-\Delta(0)) - (Q(-h) - \Delta(0)) \|_2 \\
& = \frac{1}{2} \|2 Q(h)\|_2 = \|Q(h)\|_2.
\end{align*}
In remains to take the supremum over admissible $h$ to obtain the announced lower bound.

Next, the transformation of the lower bound for the  global worst-case error
relies on a case of validity of the S-procedure due to Polyak, see \cite{polyak1998convexity}.
We start by writing this (squared) lower bound as 
{\small
$$
\inf_{\gamma \in \bR} \; \gamma 
\; \mbox{ s.to } 
\|Q(h)\|_2^2 \leq \gamma \mbox{ whenever } \|L^{1/2} h \|_2^2 \leq \eps^2, \; \|\La h \|_2^2 \leq \eta^2.
$$
}Using the S-procedure of Polyak, 
the above constraint is equivalent to the existence of $c,d\geq 0$ such that,
for all $h \in \bR^N$,
\begin{equation}
\label{cond_S}
\|Q(h)\|_2^2 - \gamma \leq c(\|L^{1/2} h \|_2^2 - \eps^2)  + d(\|\La h \|_2^2 - \eta^2),
\end{equation}
under the proviso that there exist $\wt{h} \in \bR^N$ and $\alpha,\beta \in \bR$ such that $\|L^{1/2}\wt{h}\|_2^2 < \eps^2 $,
 $\|\La \wt{h}\|_2^2 < \eta^2 $,
and $\alpha L + \beta \La^* \La \succ 0$.
This proviso is met by taking $\wt{h}=0$ and $(\alpha,\beta) = (1 - \tau, \tau)$ for any $\tau \in (0,1)$,
see \S 1 above.
Now, the constraint \eqref{cond_S} can be written as
$$
\langle (c L + d \La^* \La - Q^*Q) h, h \rangle + \gamma - c \eps^2 - d \eta^2 \ge 0
$$
for all $h \in \bR^N$,
which in fact decouples as the two constraints
 $cL + d \La^* \La  - Q^* Q \succeq 0$
and $\gamma - c \eps^2 - d \eta^2 \geq 0$.
Under the latter constraint, the minimal value of $\gamma$ is $c \eps^2 + d \eta^2$
and we arrive at the desired expression.
\qed

\vskip 5mm

\paragraph*{{\bf \S 4.} Proof of Lemma \ref{LB+UBGlobal}} 
The two additional lemmas below are needed.

\blem
If $A,B,C$ are square matrices of similar size and if $C \succeq 0$,
then
$$
\bbmx
A & \vline & 0\\
\hline
0 & \vline & B 
\ebmx 
\succeq 
\bbmx
A - C & \vline &  C\\
\hline
C & \vline & B - C 
\ebmx.
$$
\elem

\bpf
To prove that the difference of these two matrices is positive semidefinite,
we simply write, for any vectors $x,y$,
\begin{align*}
& \left\langle  
\bbmx
\; C & \vline &  -C \; \\
\hline
\; -C & \vline & C \; 
\ebmx
\bbmx x \\ \hline y \ebmx,
\bbmx x \\ \hline y \ebmx 
\right\rangle\\
& = \langle Cx,x \rangle - \langle Cy,x \rangle
- \langle Cx,y \rangle + \langle Cy,y \rangle\\
& = \langle C^{1/2}x, C^{1/2}x \rangle
- 2\langle C^{1/2}x, C^{1/2}y \rangle
+\langle C^{1/2}y, C^{1/2}y \rangle\\
& = \|C^{1/2} x - C^{1/2} y \|_2^2,
\end{align*}
which is obviously nonnegative.
\epf

\blem
If $A$ and $B$ are positive semidefinite matrices of similar size such that $A + B \succ 0$,
then $C:= A(A+B)^{-1}B$ is positive semidefinite.
\elem

\bpf
Writing $C$ as $C = A(A+B)^{-1}(A+B-A)$,
i.e., $C= A -A(A+B)^{-1}A$, shows that $C$ is self-adjoint and reveals that we in fact have to prove that $A(A+B)^{-1}A \preceq A$.
To see why this is so, we start from $A \preceq A+B$, so that
$M:= (A+B)^{-1/2}A(A+B)^{-1/2}$ satisfies $M \preceq I$.
This implies that $M^2 \preceq M$, which reads
{\small
$$
(A+B)^{-1/2}\hspace{-0.5mm}A(A+B)^{-1}A(A+B)^{-1/2} \hspace{-0.5mm}
\hspace{-1mm}\preceq \hspace{-1mm}\ (A+B)^{-1/2}\hspace{-0.5mm}A(A+B)^{-1/2}\hspace{-0.5mm}.
$$}Multiplying on the left and on the right by $(A+B)^{1/2}$ yields the desired result.
\epf

Focusing now on the proof of Lemma \ref{LB+UBGlobal},
let us consider  $c,d \ge 0$ such that 
\be 
\label{SuccQ*Q}
c L + d \La^* \La \succeq Q^* Q
\ee
and let us set $\tau = d/(c+d)$.
From \eqref{ExprReg}, we notice that
\begin{align*}
\Delta_\tau \La & = (cL + d \La^* \La)^{-1} d \La^*\La,\\
I - \Delta_\tau \La & = (cL + d \La^* \La)^{-1} c L .
\end{align*}
Multiplying \eqref{SuccQ*Q} on the right by $\bbmx I - \Delta_\tau \La \; \vline \; \Delta_\tau \La \ebmx$, which equals $(cL + d \La^* \La)^{-1} \bbmx cL \; \vline \; d \La^* \La \ebmx$,
and on the left by its adjoint, we arrive at
\begin{multline}
\label{MLT}
\bbmx cL \\ \hline d \La^* \La \ebmx
(cL + d \La^* \La)^{-1}
\bbmx cL \; \vline \; d \La^* \La \ebmx\\
 \succeq 
\bbmx (I - \Delta_\tau \La)^* \\ \hline (\Delta_\tau \La)^* \ebmx
Q^* Q 
\bbmx I - \Delta_\tau \La \;  \vline \;  \Delta_\tau \La \ebmx.
\end{multline}
First, we claim that the left-hand side of \eqref{MLT} takes the form $\bbmx A-C & \vline & C\\ \hline  C & \vline & B - C \ebmx$ with $A = cL$ and $B = d \La^* \La$.
To see this, it suffices to observe, e.g., that $A = cL$ is indeed the sum of its upper two blocks,
which is clear since these blocks are $c L (cL + d \La^* \La)^{-1} cL$
and $c L (cL + d \La^* \La)^{-1} d \La \La^*$.
Second, we claim that $C$ can be written as $C = A(A+B)^{-1}B$,
which is also clear---the relation $C = c L (cL + d \La^* \La)^{-1} d \La \La^*$ was just pointed out.
Therefore, according to our two additional lemmas,
the left-hand side of \eqref{MLT} does not exceed, 
in the positive semidefinite sense,
$\bbmx cL & \vline & 0\\ \hline 0 & \vline & d \La \La^* \ebmx$.
At this point, we have shown that
$$
\bbmx (I - \Delta_\tau \La)^* \\ \hline (\Delta_\tau \La)^* \ebmx
Q^* Q 
\bbmx I - \Delta_\tau \La \;  \vline \;  \Delta_\tau \La \ebmx
\preceq \bbmx cL & \vline & 0\\ \hline 0 & \vline & d \La^* \La \ebmx,
$$
which is equivalent to 
$$
\|Q(I-\Delta_\tau \La)f + Q(\Delta_\tau \La) g\|_2^2 \le c \|L^{1/2} f\|_2^2 + d \|\La g\|_2^2
$$ 
for all $f,g \in \bR^N$.
The observation map $\La$ is obviously surjective in the present situation\footnote{In general, it is always assumed that $\La : \bR^N \to \bR^m$ is surjective,
as it does not make sense to collect an observation that can be deduced from the others.},
so that any $e \in \bR^{n_\ell}$ can be written as $e = \La g$ for some $g \in \bR^N$. 
From here, the desired result follows.
\qed

\vskip 5mm

\paragraph*{{\bf \S 5.} Near optimality under mild overstimation of $\eps$ and $\eta$}

According to Theorem \ref{ThmGlobal} (and its proof)
and using the same notation, we have
$$
\inf_{\Delta: \bR^{n_\ell} \to \bR^n} {\rm gwce}_Q(\Delta)^2
= c_\flat \eps^2 + d_\flat \eta^2
$$
while $c_\flat L + d_\flat \La^* \La \succeq Q^* Q$.
Now suppose that $\eps$ and $\eta$ are not exactly known but overestimated by $\bar{\eps}$ and $\bar{\eta}$.
Solving the semidefinite program \eqref{SDPGlobal} with $\bar{\eps}$ and $\bar{\eta}$ provides a parameter $\bar{\tau} = \bar{d}/(\bar{c}+\bar{d})$ such that
\begin{multline*}
\sup_{\substack{\|L^{1/2} f\|_2 \le \bar{\eps} \\ \| e \|_2 \le \bar{\eta}}} \| Q(f) - Q \circ \Delta_{\bar{\tau}} (\La f + e) \|_2^2\\
= \min\left\{ c \bar{\eps}^2 + d \bar{\eta}^2: \; c L + d \La^* \La \succeq Q^* Q \right\}.
\end{multline*}
Since $\bar{\eps} \ge \eps$ and $\bar{\eta} \ge \eta$, we deduce in particular that
$$
\sup_{\substack{\|L^{1/2} f\|_2 \le \eps \\ \| e \|_2 \le \eta}} \| Q(f) - Q \circ \Delta_{\bar{\tau}} (\La f + e) \|_2^2
\le  c_\flat \bar{\eps}^2 + d_\flat \bar{\eta}^2.
$$
Under the mild overestimations $\bar{\eps} \le C \eps$ and $\bar{\eta} \le C \eta$,
this implies that 
\begin{align*}
{\rm gwce}_Q(Q \circ \Delta_{\bar{\tau}})^2
& \le C^2 \big[ c_\flat \eps^2 + d_\flat \eta^2 \big]\\
& = C^2 \inf_{\Delta: \bR^{n_\ell} \to \bR^n} {\rm gwce}_Q(\Delta)^2,
\end{align*}
proving that the recovery map $Q \circ \Delta_{\bar{\tau}}$ is globally near optimal.

\vskip 5mm

\paragraph*{{\bf \S 6.} No SDPs to optimal estimate  a linear functional}

If $Q = \langle q,\cdot \rangle : \bR^N \to \bR$ is a linear functional,
then solving the semidefinite program \eqref{SDPGlobal} 
and composing the resulting regularization map $\Delta_{\tau_\flat}$ with $Q$
to obtain a globally optimal recovery map
is quite wasteful.
In such a situation,
a globally optimal recovery map can be more directly obtained as $\langle a_\flat, \cdot \rangle$, where $a_\flat \in \bR^{n_\ell}$ is a solution to
\be
\label{ProgLF1}
\underset{a \in \bR^{n_\ell}}{\rm minimize}  \left[ \sup_{\|L^{1/2} f\|_2 \le \eps} | \langle 
q-\La^* a , f \rangle | \, \times \eps + \|a\|_2 \times \eta \right] .
\ee
This laborious-looking optimization program can be turned into a more manageable one.
For instance, if the graph $G$ has connected components $C_1,\ldots,C_K$,
then the eigenvalues of $L$ are 
$0 = \la_1 = \cdots = \la_K < \la_{K+1} \le \cdots \le \la_N$.
Denoting by $(v_1,\ldots,v_N)$ an orthonormal basis associated with these eigenvalues 
(so that $v_k = {\bm 1}_{C_k} / \sqrt{|C_k|}$, $k=1,\ldots, K$),
the problem \eqref{ProgLF1} reduces to 
\begin{align*}
& \underset{a \in \bR^{n_\ell}}{\rm minimize}   \; \left[ 
\sum_{k=K+1}^{N} \f{\langle q - \La^*a , v_k \rangle^2}{\la_k} \right]^{1/2} \hspace{-1mm} \times \eps + \|a\|_2 \times \eta\\
& \qquad \mbox{s.to } \; \langle q - \La^* a, v_k \rangle = 0, \quad k=1,\ldots,K.
\end{align*}
Note that the vector $\La^* a \in \bR^N$ appearing above is just the vector $a \in \bR^{n_\ell}$ padded with zeros on the unlabeled vertices.

\vskip 5mm

\paragraph*{{\bf \S 7.} A graph whose Laplacian is a scaled orthogonal projector}
Suppose that $G$ is an unweighted graph (so that $w_{i,j} \in \{0,1\}$ for all $i,j=1,\ldots,N$)
made of connected components $C_1,\ldots,C_K$ which are all complete graphs on an equal number~$n$ of vertices.
The adjacency matrix $W_k$ and graph Laplacian $L_k$ of each $C_k$ are
\begin{align*}
W_k & = \bbmx
0 & 1 & 1 & \cdots & 1\\
1 & 0 & 1 & \cdots & 1\\
1 & 1 & 0 & \ddots & \vdots\\
\vdots & \vdots & \ddots & \ddots & 1\\
1 & 1 & \cdots & 1 & 0
\ebmx,
\\
L_k & = \bbmx
n-1 & -1 & -1 & \cdots & -1\\
-1 & n-1 & -1 & \cdots & -1\\
-1 & -1 & n-1 & \ddots & \vdots\\
\vdots & \vdots & \ddots & \ddots & -1\\
-1 & -1 & \cdots & -1 & n-1
\ebmx .
\end{align*}
Note that $L_k$ has eigenvalue~$0$ of multiplicity $1$
and eigenvalue~$n$ of multiplicity $n-1$.
Therefore, the whole graph Laplacian 
$$
L = \bbmx
L_1 & \vline & 0 & \vline & \cdots & \vline &  0\\
\hline
0 & \vline & L_2 & \vline & \ddots & \vline & \vdots \\
\hline
\vdots & \vline & \ddots & \vline & \ddots & \vline & 0\\
\hline
0 & \vline & \cdots & \vline & 0 & \vline & L_K
\ebmx
$$
has eigenvalue~$0$ of multiplicity $K$
and eigenvalue~$n$ of multiplicity $(n-1)K$.
This means that the renormalized Laplacian $(1/n)L$ is an orthogonal projector.

\vskip 5mm

\paragraph*{{\bf \S 8.} Proof of Theorem \ref{ThmLocal}}
The argument is divided into three parts, namely:\\
{\bf a)}~there is a parameter $\tau_\natural \in (0,1)$ yielding
\be
\label{DefTauNat}
\|L^{1/2} \Delta_{\tau_\natural}(y)\|_2 = \f{\eps}{\eta} \|\La  \Delta_{\tau_\natural}(y) - y \|_2,
\ee
and the corresponding reguralizer $\Delta_{\tau_\natural}(y)$ is a solution to 
\be
\label{ProgNO}
\underset{f \in \bR^N}{\rm minimize \;} \max\Big\{  
\|L^{1/2} f\|_2^2, \f{\eps^2}{\eta^2} \|\La f - y\|_2^2\Big\};
\ee
{\bf b)}~the optimization program \eqref{ProgNO} admits a unique solution $f_\natural$ (hence equal to $\Delta_{\tau_{\natural}}(y)$);\\
{\bf c)}~the solution $f_\natural$ to \eqref{ProgNO} does provide a near optimal local worst-case error.

{\em Justification of }{\bf a)}.
For any $\tau \in [0,1]$,
let $f_\tau$ denote $\Delta_\tau(y)$.
Recalling that $f_0$ and $f_1$ are interpreted as
\begin{align*}
f_0 & = \underset{f \in \bR^N}{\argmin}  \|\La f - y\|_2 \quad \mbox{s.to } L^{1/2} f = 0,\\
f_1 & = \underset{f \in \bR^N}{\argmin} \|L^{1/2} f\|_2 \quad \mbox{s.to } \La f = y.
\end{align*}
we have 
\begin{eqnarray*}
\|L^{1/2} f_0\|_2 - \f{\eps}{\eta} \|\La f_0 - y\|_2 
& = - \df{\eps}{\eta} \|\La f_0 - y\|_2
& < 0,\\
\|L^{1/2} f_1\|_2 - \f{\eps}{\eta} \|\La f_1 - y\|_2 
& = \qquad \, \|L^{1/2} f_1\|_2  
& > 0.
\end{eqnarray*}
The continuity of $\tau \mapsto f_\tau = ((1-\tau)L + \tau \La^* \La)^{-1} (\tau \La^* y)$
guarantees that there exists some $\tau_\natural \in (0,1)$
satisfying $\|L^{1/2} f_{\tau_\natural}\|_2 - (\eps/\eta) \|\La f_{\tau_\natural} - y\|_2 = 0$, as announced in \eqref{DefTauNat}.
We additionally point out that this $\tau_\natural$ is unique,
which is a consequence of the facts that $\tau \mapsto \|L^{1/2} f_\tau \|_2$ is strictly increasing and that $\tau \mapsto \| \La f_\tau - y\|_2$ is strictly decreasing.
To see the former, say, recall that $f_\tau$ is the unique minimizer of $ ((1-\tau)/\tau) \|L^{1/2} f\|_2^2 +  \|\La f - y\|_2^2$.
Therefore, given $\sigma < \tau$,
\begin{align*}
\left( \f{1}{\sigma} - 1 \right) & \|L^{1/2} f_\sigma\|_2^2  +  \|\La f_\sigma - y\|_2^2\\
& < \left( \f{1}{\sigma} - 1 \right) \|L^{1/2} f_\tau\|_2^2 +  \|\La f_\tau - y\|_2^2\\
& = \left( \f{1}{\tau} - 1 \right) \|L^{1/2} f_\tau\|_2^2 +  \|\La f_\tau - y\|_2^2\\
& \quad
+ \left( \f{1}{\sigma} - \f{1}{\tau} \right) \|L^{1/2}f_\tau \|_2^2\\
& < \left( \f{1}{\tau} - 1 \right) \|L^{1/2} f_\sigma\|_2^2 +  \|\La f_\sigma - y\|_2^2\\
& \quad
+ \left( \f{1}{\sigma} - \f{1}{\tau} \right) \|L^{1/2}f_\tau \|_2^2.
\end{align*}
Rearranging this inequality reads 
$$
\left( \f{1}{\sigma} - \f{1}{\tau} \right)\|L^{1/2} f_\sigma \|_2^2
< \left( \f{1}{\sigma} - \f{1}{\tau} \right)\| L^{1/2} f_\tau \|_2^2,
$$
i.e., $\| L^{1/2} f_\sigma \|_2 < \| L^{1/2} f_\tau \|_2$,
as expected.
To finish, we now need to show that $f_{\tau_\natural}$ is a solution to \eqref{ProgNO}.
To this end,
we remark on the one hand that the objective function of \eqref{ProgNO} evaluated at $f_{\tau_\natural}$ is
$$
\max \left\{  
\|L^{1/2} f_{\tau_\natural}\|_2^2, \f{\eps^2}{\eta^2} \|\La f_{\tau_\natural} - y\|_2^2\right\}
= \gamma^2,
$$
where $\gamma$ is the common value of both terms in \eqref{DefTauNat}.
On the other hand,
\begin{align*}
& \mbox{setting } & 
\tau'_\natural 
& = \f{(\eta^2 / \eps^2) \tau_\natural}{1 - \tau_\natural + (\eta^2/\eps^2) \tau_\natural} \in [0,1],\\
& \mbox{so that } &  1 - \tau'_\natural 
& = \f{1-  \tau_\natural}{1 - \tau_\natural + (\eta^2 / \eps^2 ) \tau_\natural} \in [0,1],
\end{align*}
the objective function of \eqref{ProgNO}
evaluated at any $f \in \bR^N$ satisfies
{\small
\begin{align*}
& \max \left\{  
\|L^{1/2} f\|_2^2, \f{\eps^2}{\eta^2} \|\La f - y\|_2^2\right\}\\
& \ge (1-\tau'_\natural) \|L^{1/2} f\|_2^2 + \tau'_\natural \f{\eps^2}{\eta^2} \|\La f - y\|_2^2\\
& = \f{1}{1-\tau_\natural + (\eta^2/\eps^2) \tau_\natural}
\left( (1-\tau_\natural) \|L^{1/2} f\|_2^2 + \tau_\natural \|\La f - y\|_2^2 \right)\\
& \ge \f{1}{1-\tau_\natural + (\eta^2/\eps^2) \tau_\natural}
\left( (1-\tau_\natural) \|L^{1/2} f_{\tau_\natural}\|_2^2 + \tau_\natural \|\La f_{\tau_\natural} - y\|_2^2 \right)\\
& = \f{1}{1-\tau_\natural + (\eta^2/\eps^2) \tau_\natural}
\left( (1-\tau_\natural) \gamma^2 + \tau_\natural (\eta^2/\eps^2) \gamma^2 \right) = \gamma^2.
\end{align*}
}This justifies that $f_{\tau_\natural}$ is a solution to \eqref{ProgNO}.\qed

{\em Justification of }{\bf b)}.
Here, we aim at showing that \eqref{ProgNO} admits a unique minimizer.
Let $\wh{f}$ and $\mu^2$ denote a minimizer and the minimal value of \eqref{ProgNO}, respectively.
We first claim that 
\be
\label{EqatWHF}
\|L^{1/2} \wh{f}\|_2 = \f{\eps}{\eta} \|\La \wh{f} - y \|_2 = \mu .
\ee
Indeed, suppose e.g. that $\|L^{1/2} \wh{f} \|_2 < (\eps/\eta) \|\La \wh{f}-y\|_2 =~\mu$.
Pick an $h \in \bR^N$ such that $\langle \La \wh{f} - y, \La h \rangle \not= 0$
(which exists,~for otherwise $\La^*(\La \wh{f}-y)=0$, hence $\La \wh{f}-y = \La \La^*(\La \wh{f} -y) =~0$, and so $\mu = 0$, in which case $\|L^{1/2} \wh{f} \|_2 < \mu$ cannot occur).
Then, considering $\wh{f}_t := \wh{f} + t h$ for a small enough $t$ in absolute value,
we see that 
$$
\f{\eps}{\eta} \|\La \wh{f}_t - y\|_2
= \f{\eps}{\eta} \left(  
\|\La \wh{f} - y\|_2  + t \langle \La \wh{f} - y, \La h \rangle + o(t)
\right)
$$
can be made smaller that $\mu$,
while $\|L^{1/2} \wh{f}_t \|_2$ can remain smaller than $\mu$.
This contradicts the defining property of $\wh{f}$
and establishes \eqref{EqatWHF}.

Now let $\wh{f}$ and $\wt{f}$ be two minimizers of~\eqref{ProgNO}.
Applying \eqref{EqatWHF} to $\wh{f}$, $\wt{f}$, and  $(\wh{f}+\wt{f})/2$,
which is also a minimizer of \eqref{ProgNO}, yields
{\small
\begin{align*}
\bigg\| \f{1}{2}L^{1/2}\wh{f} + \f{1}{2} L^{1/2}\wt{f}  \bigg\|_2
& =   \big\|L^{1/2} \wh{f} \big\|_2 = \big\|L^{1/2} \wt{f} \big\|_2 = \mu,\\
\bigg\| \f{1}{2}  \La ( \wt{f} - y ) + \f{1}{2}  \La ( \wh{f} - y ) \bigg\|_2
& = \big\| \La \wh{f} - y \big\|_2 = \big\| \La \wt{f} - y \big\|_2 = \f{\eta}{\eps} \mu,
\end{align*}
}which forces $L^{1/2} \wh{f} = L^{1/2} \wt{f}$ and $\La \wh{f} = \La \wt{f}$, implying that
i.e. $\wh{f} - \wt{f} \in \ker(L^{1/2}) \cap \ker (\La) =  \ker(L) \cap \ker (\La) = \{0\}$,
i.e., that $\wh{f} = \wt{f}$ is a unique minimizer.\qed

{\em Justification of }{\bf c)}.
Since the original signal $f \in \bR^N$ that we  try to recover satisfies $\|L^{1/2} f\|_2 \le \eps$
and $\|\La f - y\|_2 \le \eta$,
it is clear that the minimizer $\wh{f}$ of \eqref{ProgNO} satisfies $\|L^{1/2} \wh{f}\|_2 \le \eps$
and $\|\La \wh{f} - y\|_2 \le \eta$, too.
In other words, it is model- and data-consistent,
which always leads to near optimality of the local worst case error with a factor $2$.
Indeed,
considering the set $\{Q(f) \colon f \in \cK, \, e \in \cE, \, \La f + e = y\}$,
let $f^\star$ denote its Chebyshev center (in our situation, it exists and is unique, see \cite{garkavi1962optimal}).
Then, for any $f \in \cK$ and $e \in \cE$ with $\La f + e = y$,
we have
\begin{align*}
\|Q(f) - Q(\wh{f}) \|_2
& \le \|Q(f) - Q(f^\star) \|_2 + \|Q(\wh{f}) - Q(f^\star) \|_2\\
& \le {\rm lwce}_Q(y,f^\star) + {\rm lwce}_Q(y,f^\star)\\
& = 2 \, \inf_{z \in \bR^N} {\rm lwce}_Q(y,z).
\end{align*}
Taking the supremun over the admissable $f \in \cK$ and $e \in \cE$ gives ${\rm lwce}_Q(y,\wh{f}) \le 2 \, \inf_{z \in \bR^N} {\rm lwce}_Q(y,z)$,
as desired.\qed

\vskip 5mm

\begin{figure}[!htbp]
\centering
\subfigure[]{\includegraphics[width=0.24\textwidth]{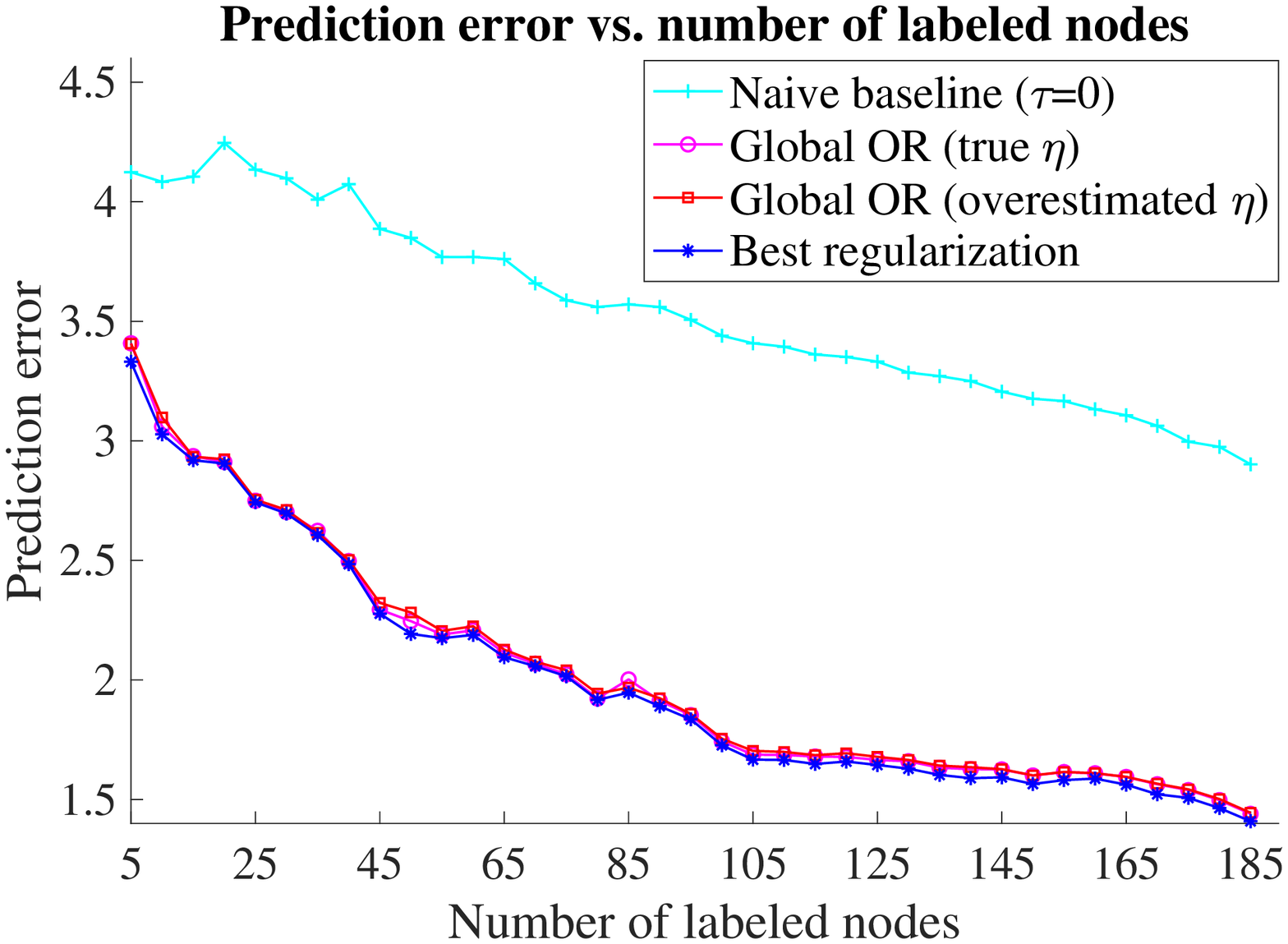}}
\subfigure[]{\includegraphics[width=0.24\textwidth]{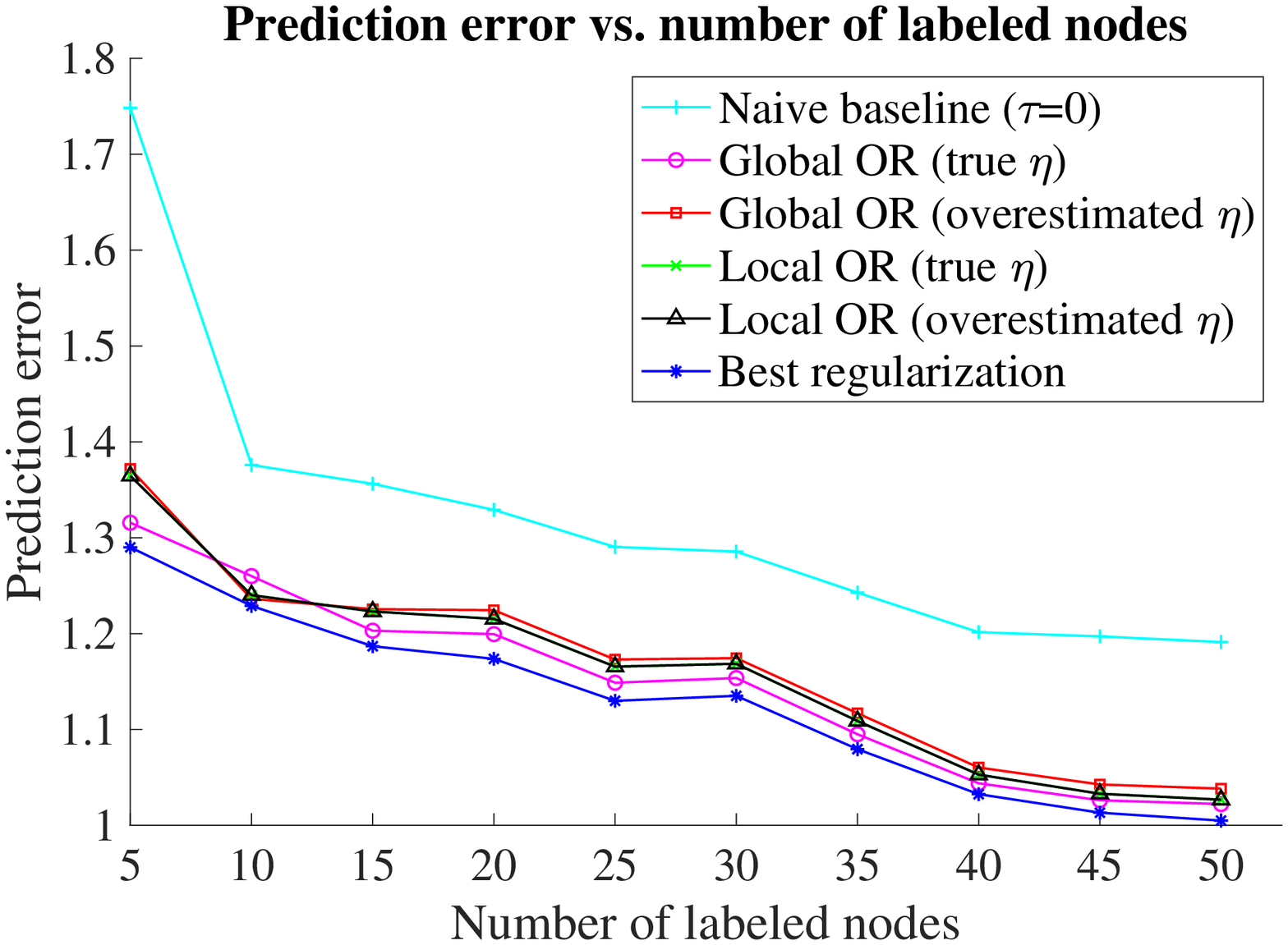}}
\caption{Prediction errors vs. number of labeled nodes on two different graphs with additive noise proportional to degree: (a) Netscience (b) polbooks.}
\label{pred_error_model3}
\end{figure}

\begin{figure}[!htbp]
\centering
\subfigure[]{\includegraphics[width=0.24\textwidth]{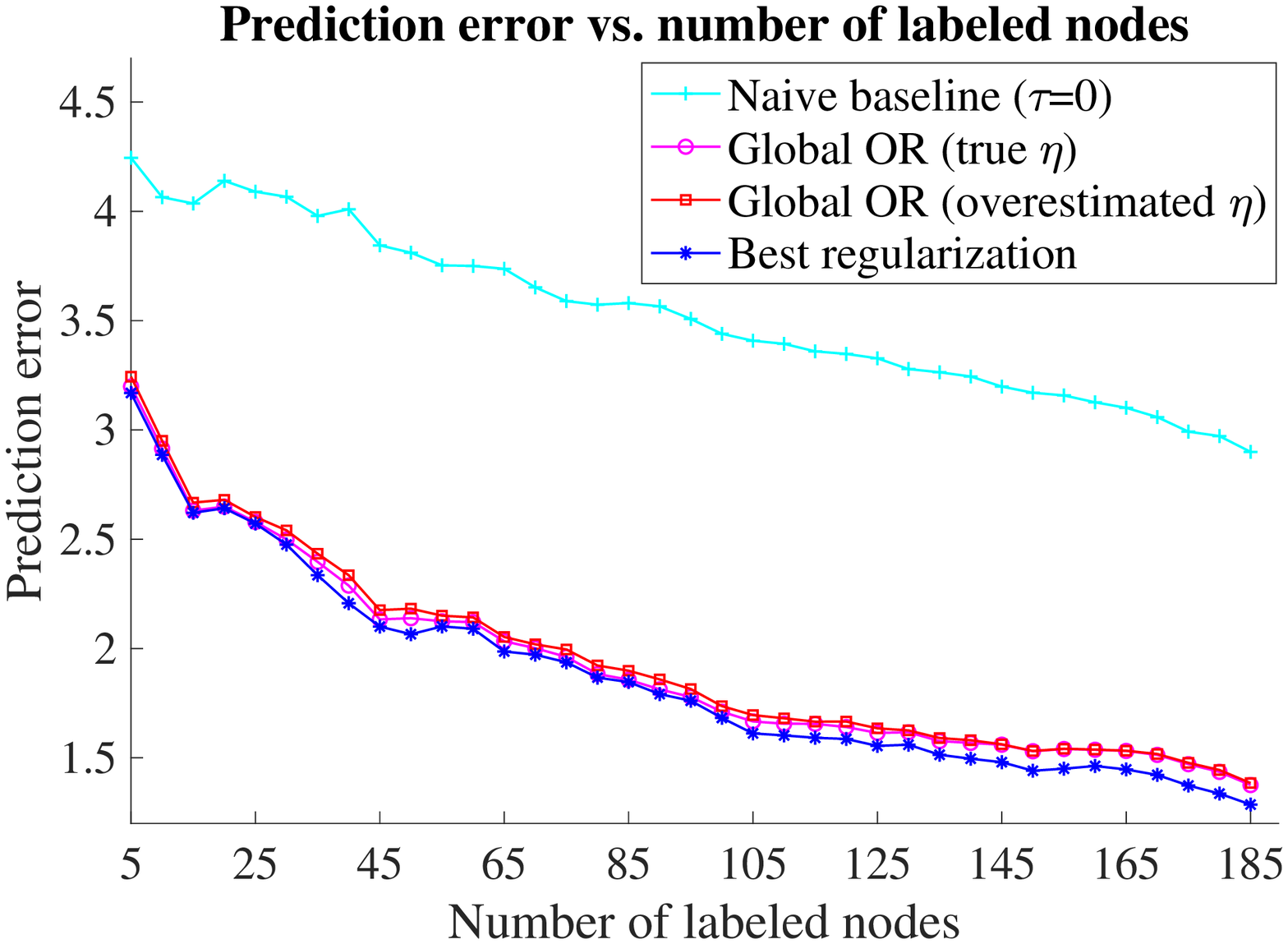}}
\subfigure[]{\includegraphics[width=0.24\textwidth]{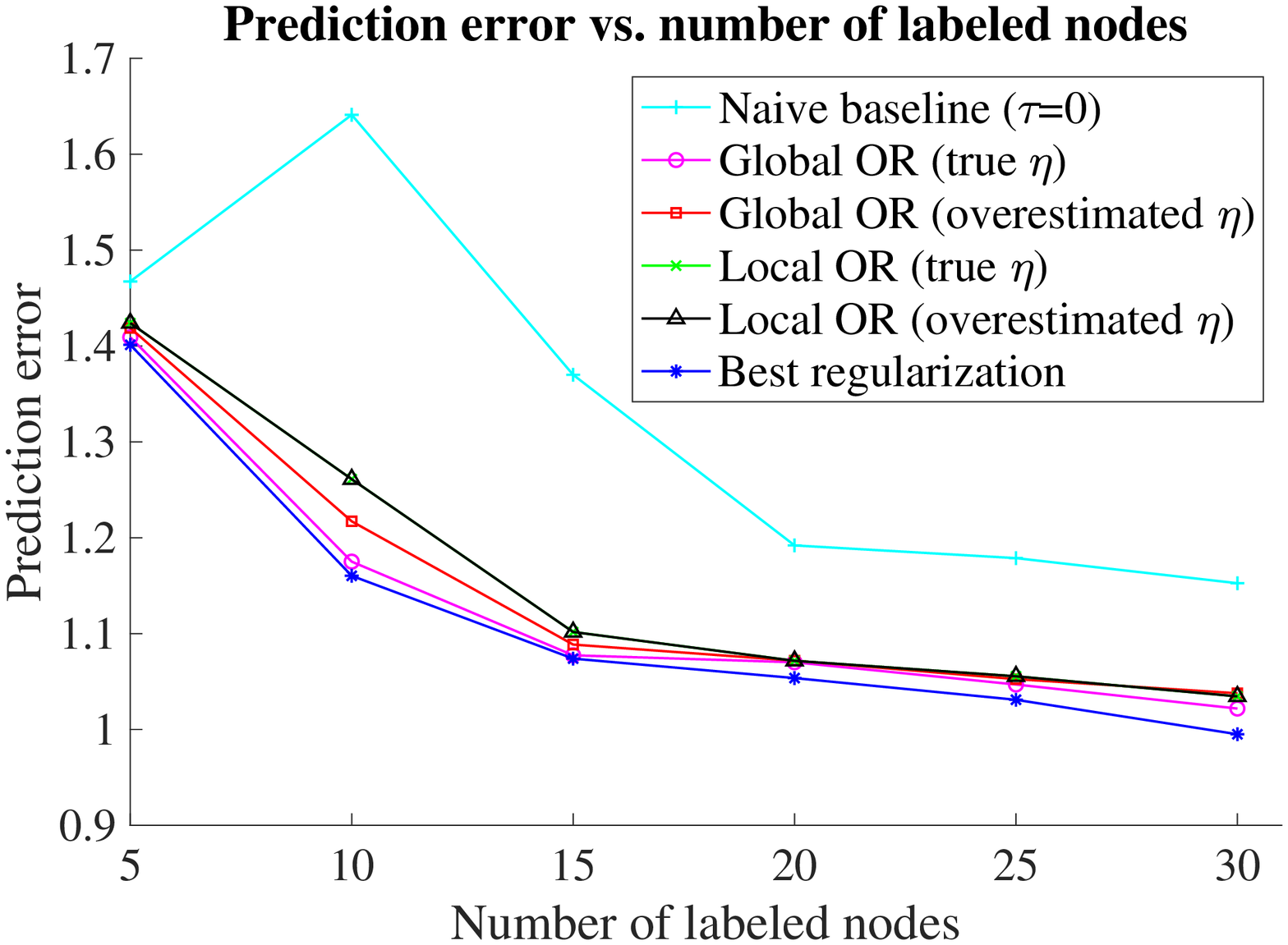}} 
\caption{Prediction errors vs. number of labeled nodes on two different graphs with additive noise proportional to the inverse of degree: (a) Netscience (b)~dolphins.}
\label{pred_error_model4}
\end{figure}

\paragraph*{{\bf \S 9.} Implementation details and additional experiments} 
We consider the following well-known graph datasets: adjnoun (112 nodes, 425 edges)~\citep{newman2006finding}, 
Netscience (379 nodes, 914 edges)~\citep{girvan2002community}, 
polbooks (105 nodes, 441 edges)~\citep{polbooks}, lesmis (77 nodes, 254 edges)~\citep{}, 
and dolphins (62 nodes, 159 edges)~\citep{lusseau2003bottlenose}. 
All of these can be downloaded from the Suitesparse Matrix Collection~\citep{suitesparse2011davis}. 
When generating synthetic signals, 
we follow an approach similar to Equation (15) in~\citep{dong2019graph}. 
Let $L = \chi D \chi^T$ be an eigendecomposition of the graph Laplacian, 
let $D^\dagger$ be the pseudoinverse of $D$,
and let $c \sim \mathcal{N}(0, D^\dagger) $ be a Gaussian vector. 
The ground truth labels are then given by $f = \chi c$. 
The main difference with~\citep{dong2019graph} is that
 $f$ is not assumed to be corrupted simply by Gaussian noise, 
 but we consider different additive noise vectors satisfying $\|e \|_2 \leq \eta$. 
 In the main text, the plots were shown for a noise vector that is generated by taking a uniform random noise vector and subtracting the mean, 
 and before scaling to ensure that $\|e \|_2 \leq \eta$. 
 Here, to illustrate results of a more deterministic flavor, 
 we show results for noise of magnitude proportional to the node degree (Figure~\ref{pred_error_model3}) and to the inverse degree (Figure~\ref{pred_error_model4}).

\begin{figure}[!htbp]
\centering
\subfigure[]{\includegraphics[width=0.24\textwidth]{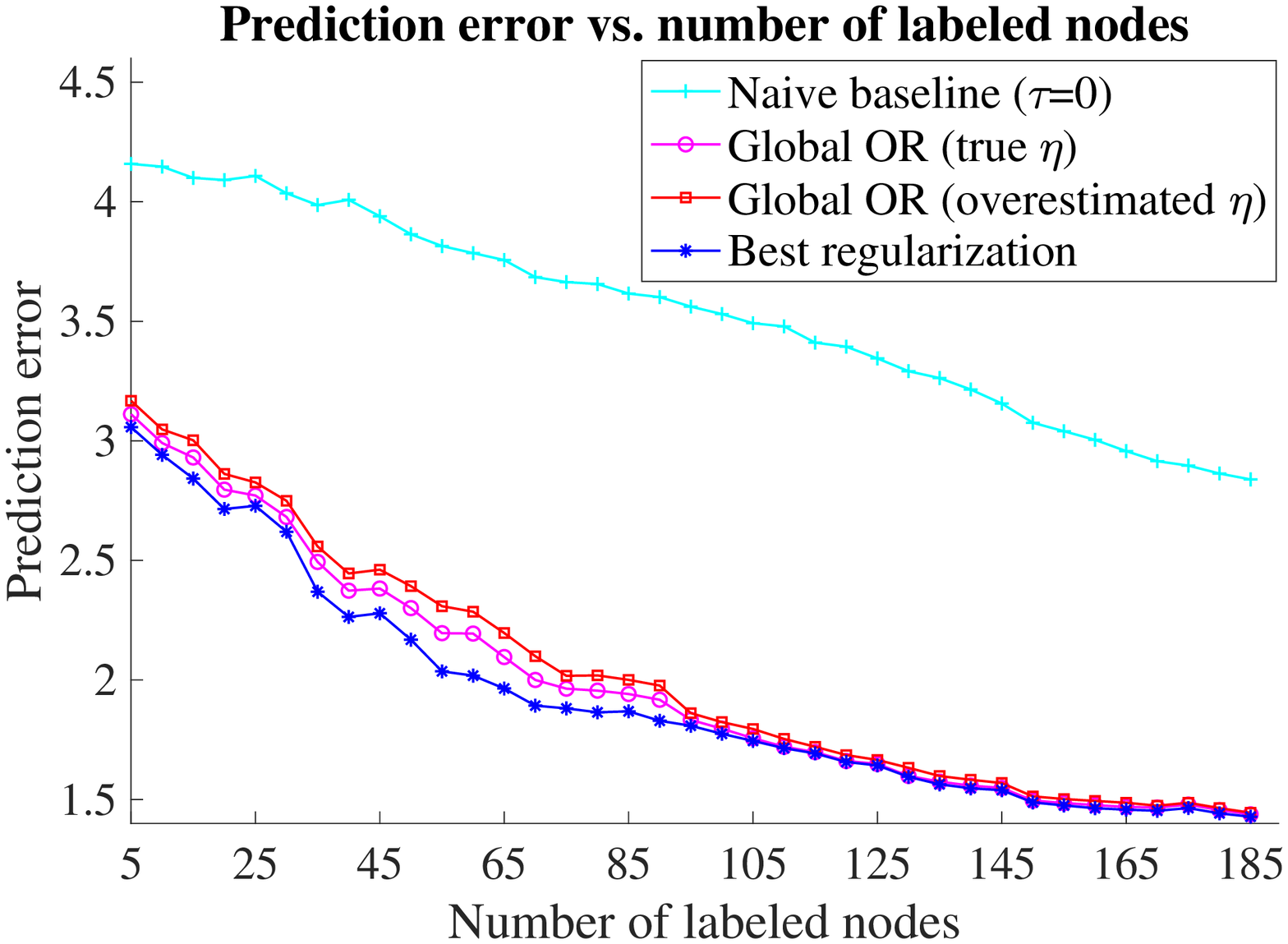}}
\subfigure[]{\includegraphics[width=0.24\textwidth]{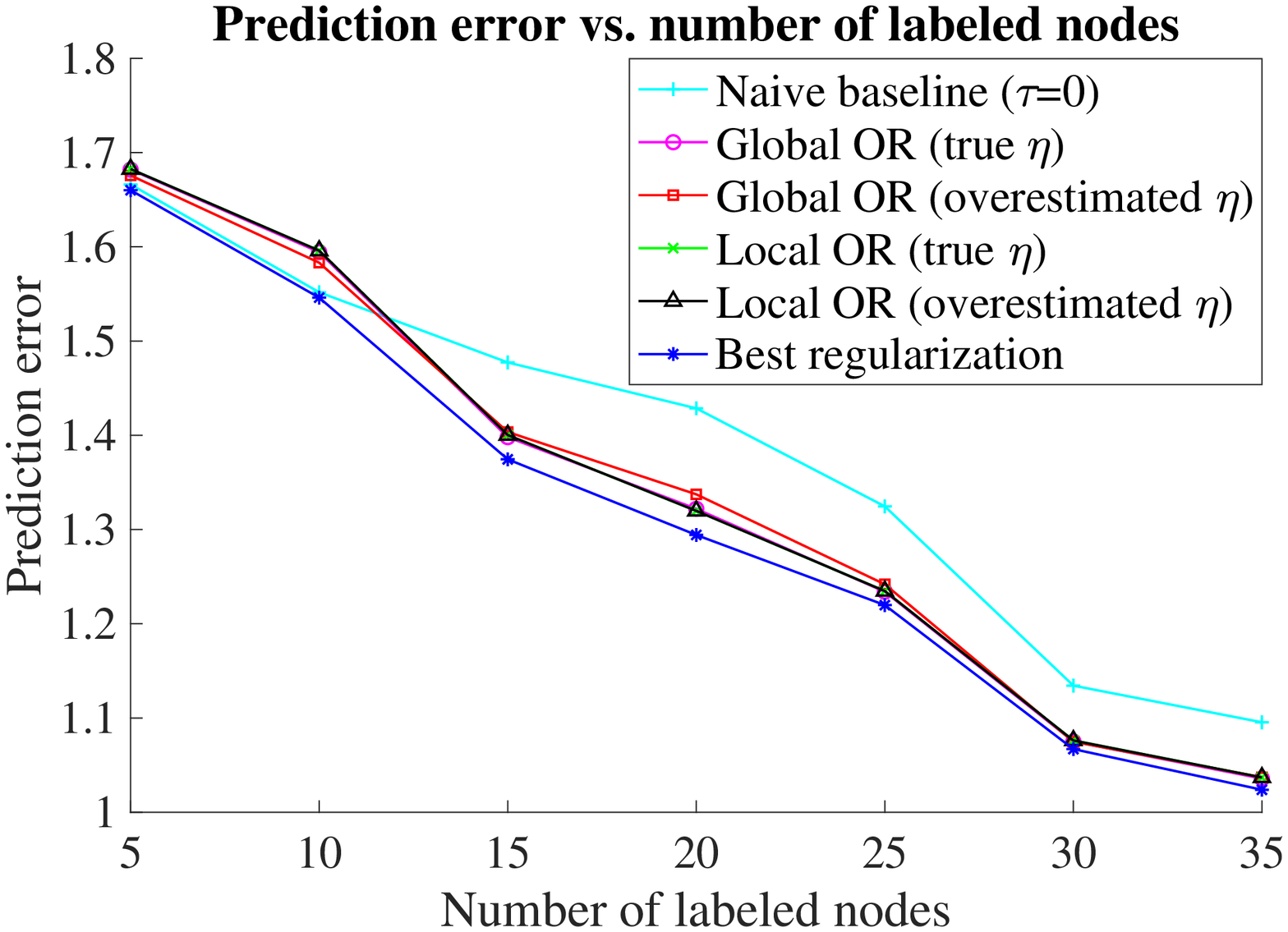}}
\caption{Prediction errors vs. number of labeled nodes on two different graphs with additive noise generated uniformly: (a) Netscience (b) lesmis.}
\label{pred_error_model1}
\end{figure}

Keeping the same parameters as those used in the main text,
we test several optimal recovery methods on different graphs and with different error models.
Figures~\ref{pred_error_model1}-\ref{pred_error_model4} support our conclusions that a mild overestimation of $\eta$ does not lead to bad prediction error 
and that the prediction errors attached to the locally/globally optimal recovery maps are close to the smallest prediction error possible for any choice of regularization parameter. 
This confirms that these methods provide a suitable way to choose regularization parameters. 

It is worth pointing out that the globally optimal recovery map is linear since the regularization parameter does not depend on the observation vector $y$.
In contrast,
the locally near optimal recovery map is nonlinear since the unique parameter $\tau_\natural$ satisfying 
$$
\|L^{1/2} \Delta_\tau(y)\|_2 = \f{\eps}{\eta} \|\La  \Delta_\tau(y) - y \|_2
$$
does depend on the observation vector $y$. 
When implementing globally optimal recovery maps, we compute the globally optimal regularization parameter $\tau_\flat$ for each $n_\ell$ once and make a prediction when receiving different observation vectors $y$. 
For locally optimal recovery maps, we have to recompute the locally near optimal parameter $\tau_\natural$ when receiving new observation vectors $y$. 
Therefore, it is recommended to opt for the globally optimal recovery map in order to reduce computational complexity, see e.g. Figures~\ref{pred_error_model1}(a), \ref{pred_error_model3}(a), and \ref{pred_error_model4}(a)
where the locally near optimal recovery map 
is not executed.
However, dealing with large graphs may result
in semidefinite programs that cannot be run, 
so it can be better to implement the locally near optimal recovery map by using the bisection method to find the near optimal parameter $\tau_\natural$.

\vskip 5mm

\paragraph*{{\bf \S 10.} Numerical computation of the upper bound} 
Given $y \in \bR^{n_\ell}$,
the square of the local worst-case error ${\rm lwce}_Q(y,z)$ for the estimation of $Q$ by $z\in\bR^n$ is 
$$ 
\sup_{f\in\bR^N} \|Q(f)-z\|_2^2 \quad \mbox{ s.to } \|L^{1/2}f\|_2^2\leq\epsilon^2, \|\Lambda f-y\|_2^2\leq\eta^2.
$$
Introducing a slack variable $\gamma$,
we write the above optimization program as 
\begin{align*}
\inf_{\gamma} \gamma \quad &\mbox{ s.to } \|Q(f)-z\|_2^2\leq\gamma \\ 
&\mbox{ whenever } \|L^{1/2}f\|_2^2\leq\epsilon^2, \|\Lambda f-y\|_2^2\leq\eta^2.
\end{align*}
The constraint is a consequence of (but is not equivalent to) the existence of $c,d\geq0$ such that
\begin{equation*}
\|Q(f)-z\|_2^2 - \gamma \leq c(\|L^{1/2}f\|_2^2-\epsilon^2) + d(\|\Lambda f-y\|_2^2 - \eta^2)
\end{equation*}
for all $f\in\bR^N$.
The latter can be also reformulated as the condition that, for all $f\in\bR^N$,
\begin{multline*}
\left\langle(cL+d\Lambda^*\Lambda-Q^*Q)f,f\right\rangle - 2\left\langle Q^*z-\Lambda^*y,f \right\rangle \\
 + \gamma - \|z\|_2^2 - c\epsilon^2+d(\|y\|_2^2-\eta^2) \geq 0,  
\end{multline*}
or, more succinctly, that
{\footnotesize
\be
\label{SDC4UB}
\bbmx cL+d\La^*\La - Q^*Q & \vline & Q^*z-d\La^*y\\ \hline (Q^*z-d\La^*y)^* & \vline & \gamma - \|z\|_2^2 - c\epsilon^2+d(\|y\|_2^2-\eta^2) \ebmx \succeq 0.
\ee
}We conclude from the above considerations that the squared local worst-case error is upper-bounded by the optimal value of a semidefinite program, namely
\begin{align*}
&{\rm lwce}_Q(y,z)^2 \leq \inf_{\substack{\gamma \\ c,d\geq0}} \gamma \quad \mbox{ s.to }  \mbox{\eqref{SDC4UB}}.
\end{align*}

\end{document}